%% file: eacl2023.tex
\pdfoutput=1
\documentclass[11pt]{article}

\usepackage[]{EACL2023}

\usepackage{times}
\usepackage{latexsym}
\usepackage{multirow}
\usepackage{booktabs}
\usepackage{float}
\usepackage{graphics}

\usepackage[T1]{fontenc}
\usepackage[utf8]{inputenc}

\usepackage{microtype}

\usepackage{inconsolata}

\title{Exploring the Use of Foundation Models for Named Entity Recognition\\ and Lemmatization Tasks in Slavic Languages}

\author{Gabriela Pałka \and Artur Nowakowski\thanks{$\>$ Artur Nowakowski is a scholarship recipient of the Adam Mickiewicz University Foundation for the 2022/2023 academic year.} \\
        Faculty of Mathematics and Computer Science, Adam Mickiewicz University, Poznań, Poland \\
        \texttt{\{gabriela.palka,artur.nowakowski\}@amu.edu.pl} \\}

\begin{document}
\maketitle
\begin{abstract}
% Very funny abstract about grumpy cats, unicorns and Gabis.
This paper describes Adam Mickiewicz University's (AMU) solution for the 4th Shared Task on SlavNER. The task involves the identification, categorization, and lemmatization of named entities in Slavic languages. Our approach involved exploring the use of foundation models for these tasks. In particular, we used models based on the popular BERT and T5 model architectures. Additionally, we used external datasets to further improve the quality of our models. Our solution obtained promising results, achieving high metrics scores in both tasks. We describe our approach and the results of our experiments in detail, showing that the method is effective for NER and lemmatization in Slavic languages. Additionally, our models for lemmatization will be available at: \url{https://huggingface.co/amu-cai}.
\end{abstract}

\section{Introduction}
Named entity recognition and lemmatization are important tasks in natural language processing. Fine-tuning pre-trained neural language models has become a popular approach to achieve the best results in these tasks. However, the performance of this method can vary across languages and language families. In this paper, we investigate the performance of fine-tuned, language-specific neural language models in named entity recognition and lemmatization in a set of Slavic languages and compare them with multilingual solutions. 

We describe Adam Mickiewicz University's (AMU) solution for the 4th Shared Task on SlavNER, which is a part of The 9th Workshop on Slavic Natural Language Processing (Slavic NLP 2023).
Our solution is based on foundation models~\cite{foundation-models}. In particular, we used models based on the popular BERT and T5 model architectures. To increase the effectiveness of our approach, we conducted experiments with different versions of monolingual and multilingual models, investigating the potential benefits of each model variant for specific tasks.
The data provided by the organizers and external resources used for named entity recognition and lemmatization were processed and prepared as described in section~\ref{data}. Specific details regarding the approach are further discussed in section~\ref{approach}.

In order to evaluate the effectiveness of our method, we performed several experiments on the previous Shared Task edition test set. This particular set was chosen because it is a well-known benchmark for named entity recognition and lemmatization in Slavic languages. The results of our experiments are described in section~\ref{results}.

\section{Data}
\label{data}
This section provides a brief description of the datasets used in our solution. In addition to the data released by the organizers, we also used external datasets for named entity recognition and lemmatization. All training and validation samples containing named entities were converted to a CoNLL-2003 dataset format \cite{tjong-kim-sang-de-meulder-2003-introduction}.

\subsection{Shared Task Dataset}
The 4th Shared Task on SlavNER focuses on recognition, lemmatization, and cross-lingual linking of named entities in Polish, Czech and Russian languages.
The training and validation data provided by the organizers come from the previous editions of the Shared Task and consist of news articles related to a single entity or event such as Asia Bibi, Brexit, Ryanair, Nord Stream, COVID-19 pandemic and USA 2020 Elections.
The documents contain annotations of the following named entities: person (PER), location (LOC), organization (ORG), event (EVT) and product (PRO)~\cite{piskorski-etal-2021-slav}.

To obtain NER training and validation samples in the CoNLL-2003 format, we processed the data using the code provided by the Tilde team~\cite{tilde}\footnote{\url{https://github.com/tilde-nlp/BSNLP_2021}}.

\subsection{External NER Datasets}
\label{ner-data}
One way to improve the performance of NER models is to use external NER datasets to increase the volume of the training data.
These datasets contain pre-labeled documents that have been annotated with named entities, and can be used to fine-tune existing models.
This technique allows the model to learn from the additional data, which can provide a more comprehensive understanding of the context and complexities of the named entities.

\subsubsection{Collection3}
The \textit{Collection3} dataset~\cite{collection3} is based on \textit{Persons-1000}, a publicly available Russian document collection consisting of 1,000 news articles.
Currently, the dataset contains 26,000 annotated named entities (11,000 persons, 7,000 locations and 8,000 organizations).

\subsubsection{MultiNERD}
The \textit{MultiNERD} dataset~\cite{tedeschi-navigli-2022-multinerd} covers 10 languages, including Polish and Russian, and contains annotations of multiple NER categories, from which we extracted categories present in the Shared Task.
The labels were obtained by processing the Wikipedia and Wikinews articles.
In addition, the sentences were tagged automatically, in a way that can also be adapted to the Czech language.

\subsubsection{Polyglot-NER}
A \textit{Polyglot-NER} dataset~\cite{polyglotner} covers 40 languages, including Polish, Czech and Russian.
The annotations were automatically generated from Wikipedia and Freebase.
The obtained entity categories are: person, location and organization.

\subsubsection{WikiNEuRal}
The \textit{WikiNEuRal} dataset~\cite{tedeschi-etal-2021-wikineural-combined} consists of named entities in the following categories: person, location, organization and miscellaneous.
Wikipedia was used as the source for the labels, which were automatically obtained using a combination of knowledge-based approaches and neural models.
The datasets cover 9 languages, including Polish and Russian.

\subsection{External Lemmatization Datasets}
Lemmatization, the process of reducing a word or phrase to its base form, is an essential component, especially for tasks such as information retrieval and text mining.
External lemmatization datasets can improve the quality of lemmatization models by providing additional training samples that contain more inflectional variants of phrases.
Such data consists of inflected words, collocations or phrases with corresponding lemmatized forms.

\subsubsection{SEJF}
\textit{SEJF}~\cite{sejf} is a linguistic resource consisting of a grammatical lexicon of Polish multi-word expressions.
It contains two modules: an intensional module, which consists of 4,700 multiword lemmas assigned to 100 inflection graphs, and an extensional module, which contains 88,000 automatically generated inflected forms annotated with grammatical tags.

\subsubsection{SEJFEK}
\textit{SEJFEK}~\cite{savary-etal-2012-sejfek} refers to a lexical and grammatical resource related to Polish economic terms.
It contains a grammatical lexicon module with over 11,000 terminological multi-word units and a fully lexicalized shallow grammar with over 146,000 inflected forms, which was produced by an automatic conversion of the lexicon.

\subsubsection{PolEval 2019: Task 2}
\textit{PolEval 2019: Task 2}~\cite{poleval2019} is a part of a workshop focusing on natural language processing in the Polish language.
The main goal of this task was to lemmatize proper names and multi-word phrases. The train set consists of over 24,000 annotated and lemmatized phrases.
The validation set and the test set contain 200 and 1,997 phrases, respectively. 

\subsubsection{Machine Translation of External Datasets}
Due to the lack of external Czech and Russian datasets dedicated to lemmatization tasks, we decided to use OPUS-MT~\cite{tiedemann-thottingal-2020-opus}, which is a resource containing open-source machine translation models.
We machine translated all the samples prepared from the three aforementioned datasets.

\section{Approach}
\label{approach}
We participated in the two subtasks of the Multilingual Named Entity Recognition Task - \textit{Named Entity Mention Detection and Classification} and \textit{Named Entity Lemmatization}.
The solution involved fine-tuning the foundation models using task-specific modifications and additional training data. All models used in the experiments can be found on the Hugging Face Hub\footnote{\url{https://huggingface.co/models}}.  

\subsection{Named Entity Recognition}
Recently, the BERT~\cite{devlin-etal-2019-bert} model architecture has been adapted to address Slavic languages such as Polish, Czech and Russian, among others.
These languages present unique challenges because of their complex grammatical structures, declensions and inflections, making NLP tasks even more difficult. However, the application of BERT to these languages has resulted in significant improvements in language processing and understanding.

In our solution, we used several monolingual BERT models to better handle the specific linguistic nuances of individual Slavic languages. In particular, we employed of the following models: HerBERT~\cite{mroczkowski-etal-2021-herbert} for Polish, Czert~\cite{sido-etal-2021-czert} for Czech and RuBERT~\cite{rubert} for Russian.
For comparison, we also used multilingual BERT models that can handle multiple languages, including Slavic BERT~\cite{slavic-bert} and XLM-RoBERTa~\cite{xlm-roberta}. 

In the experiments, we also added a Conditional Random Fields (CRF) layer on the top of each BERT model.
A similar approach of combining CRF with neural networks has been used previously~\cite{lample-etal-2016-neural-crf}, as the CRF layer can capture the dependencies between neighboring tokens and provide a smoother transition between different entity types.

\subsection{Lemmatization}
Models based on the T5~\cite{2020t5} model architecture have achieved state-of-the-art results in various natural language processing challenges and can be fine-tuned for specific tasks.
One of the applications of T5 can be lemmatization, the process of reducing a word or phrase to its basic form (lemma).
In Slavic languages such as Polish, Czech and Russian, lemmatization is particularly important due to the complex inflection of these languages.

We approached the lemmatization task as a text-to-text problem. The input to the model is an inflected phrase or named entity, which can consist of several word forms. For example, it can consist of nouns in singular or plural form, or verbs in different tenses. The output of the model is the base, normalized form of the phrase or named entity.

To address the lack of dedicated models for Czech and Russian, we used one monolingual and a multilingual T5 model. Specifically, we chose plT5~\cite{plT5} for Polish and mT5~\cite{xue-etal-2021-mt5} for multilingual experiments. For comparison purposes, we also conducted our experiments on the small, base and large sizes of the above models. 

In the multilingual experiments, we included a language token (>>pl<<, >>cs<<, >>ru<<) as the first token of the source phrases, depending on the language of the phrase. Our preliminary experiments have shown that incorporating the language token improves the results, increasing the exact match by approximately 2 points in each language. We noticed that the model sometimes tends to change the grammatical number from plural to singular - possibly due to the fact that singular named entities occur more often in the training data.

\section{Results}
\label{results}

\begin{table*}[h!]
\centering
\include{table_val_ner}
\caption{Results of case-sensitive F1 score for named entity recognition on the COVID-19 and USA 2020 Elections test sets from the 3rd Shared Task on SlavNER. For each language in a given test set, the best score for the monolingual and multilingual solution is shown in bold. In addition, the best score for each language in a given test set is underlined.}
\label{ner-val-results}
\end{table*}

\begin{table*}[h!]
\centering
\small
\include{table_val_lemma}
\caption{Results of the case-insensitive exact match for lemmatization on the COVID-19 and USA 2020 Elections test sets from the 3rd Shared Task on SlavNER. For each test set, the best score in a given language is shown in bold and underlined.}
\label{lemma-val-results}
\end{table*}

\subsection{Named Entity Recognition Results}
The results of our named entity recognition experiments are presented in table~\ref{ner-val-results}. We evaluated our models with a case-sensitive F1 score, which is a standard span-level metric calculated on the ConLL-2003 dataset format. As test sets, we choose COVID-19 and USA 2020 Elections subsets of the 3rd Shared Task on SlavNER. 

We tested our solution in two approaches: monolingual and multilingual. For Polish and Czech, we found that monolingual models perform better for language-specific data.
In the case of Russian, multilingual models strongly outperform language-specific solutions. We assume that this is due to the lack of sufficient data for this language. In addition, multilingual models can learn common rules in Slavic languages to overcome weaknesses related to insufficient data.

We also found that adding a CRF layer significantly improves the quality of the models in most cases. However, including external datasets worsens the results in almost all cases. We suspect that this is due to the specific domain of the test sets, which are news articles. In addition, some annotation errors can be found in all datasets presented in the~\ref{ner-data} section.

\begin{table*}[h!]
\centering
\include{table_test}
\caption{Results of our systems on the released test set for named entity recognition and normalization (lemmatization). The scores are computed as case-insensitive strict matching for recognition and case-insensitive F1 score for normalization. All scores were received from the organizers.}
\label{test-results}
\end{table*}

\subsection{Lemmatization Results}
The results of our lemmatization experiments are presented in the table~\ref{lemma-val-results}. We evaluated our models with a case-insensitive exact match on the same test sets as for named entity recognition, but only on the data specific to this task. 

We tested our solution based on two models: a monolingual plT5 (only for the Polish language), and a multilingual mT5 model. We observed that the addition of each external dataset significantly improves the quality of the Polish language-specific model. Moreover, the addition of the data from PolEval 2019 also improves the results for the multilingual model.
Unfortunately, the addition of data from the lexicon generated by machine translation of the SEJF and SEJFEK datasets causes a decrease in the model performance for the Czech and Russian languages. We assume that this is due to the quality of the translation of the phrases into these languages. 

We also noticed that the quality of the lemmatization improves as the size of the model increases in almost all cases. However, for Polish, the small model trained on all available data is better than the base model. Furthermore, it is only 3 points worse than the large model, so it can be used efficiently considering the hardware limitations. 

\subsection{The 4th Shared Task on SlavNER Results}
The current edition of the shared task features news articles about the Russian-Ukrainian war, and the test set includes raw texts in Polish, Czech and Russian languages. 

As a solution, we submitted four systems consisting of the following fine-tuned models with an additional CRF layer for named entity recognition:
\begin{itemize}
    \item System 1: HerBERT\textsubscript{LARGE} for Polish trained on all available data, Czert for Czech and RuBERT for Russian trained only on the data provided by the organizers,
    \item System 2: XLM-RoBERTa\textsubscript{LARGE} for all languages trained only on the data provided by the organizers,
    \item System 3: XLM-RoBERTa\textsubscript{LARGE} for all languages trained on all available data,
    \item System 4: HerBERT\textsubscript{LARGE} for Polish, Czert for Czech and RuBERT for Russian trained on all available data.
\end{itemize}
In all the systems mentioned above, we used the following lemmatization models: plT5\textsubscript{LARGE} for Polish (trained on all available data) and mT5\textsubscript{LARGE} for Czech and Russian (trained on the data provided by the organizers and the data from PolEval 2019 Task 2).

The best solution for recognizing and categorizing named entities turned out to be System 2, which also achieved the best results for normalization (lemmatization). In addition, the normalization scores are highly dependent on the NER results, since only recognized entities are normalized.

\section{Conclusions}
We described the Adam Mickiewicz University's (AMU) participation in the 4th Shared Task on SlavNER for named entity recognition and lemmatization tasks. Our experiments encompassed various foundation models, including monolingual and multilingual BERT and T5 models. We found that incorporating a CRF layer enhanced the quality of our named entity recognition models. Additionally, our results indicate that the use of T5 models for lemmatization yields high-quality lemmatization of named entities. We will release the lemmatization models to the community and make them available at: \url{https://huggingface.co/amu-cai}.

% Entries for the entire Anthology, followed by custom entries
\bibliography{bibliography}
\bibliographystyle{acl_natbib}

\end{document}

%% file: table_val_ner.tex
\resizebox{\linewidth}{!}{%
\begin{tabular}{l|rrrrrr|rrrrrr}
\toprule
\multicolumn{1}{c|}{\multirow{3}{*}{Model}} & \multicolumn{6}{c|}{original data}                                                                                       & \multicolumn{6}{c}{+ external datasets}                                                      \\
\multicolumn{1}{c|}{}                       & \multicolumn{3}{c|}{COVID-19}                                         & \multicolumn{3}{c|}{USA 2020 Elections}          & \multicolumn{3}{c|}{COVID-19}                       & \multicolumn{3}{c}{USA 2020 Elections} \\
\multicolumn{1}{c|}{}                       & pl             & cs             & \multicolumn{1}{r|}{ru}             & pl             & cs             & ru             & pl             & cs    & \multicolumn{1}{r|}{ru}    & pl          & cs          & ru         \\
\midrule
HerBERT\textsubscript{BASE}                                & 79.50          & -              & \multicolumn{1}{r|}{-}              & 89.27          & -              & -              & 78.70          & -     & \multicolumn{1}{r|}{-}     & 84.63       & -           & -          \\
HerBERT\textsubscript{BASE} + CRF                          & 80.11          & -              & \multicolumn{1}{r|}{-}              & 90.16          & -              & -              & 80.86          & -     & \multicolumn{1}{r|}{-}     & 87.43       & -           & -          \\
HerBERT\textsubscript{LARGE}                               & 81.18          & -              & \multicolumn{1}{r|}{-}              & 91.71          & -              & -              & 81.29          & -     & \multicolumn{1}{r|}{-}     & 89.83       & -           & -          \\
HerBERT\textsubscript{LARGE} + CRF                         & 81.75          & -              & \multicolumn{1}{r|}{-}              & \underline{\textbf{92.13}} & -              & -              & \underline{\textbf{82.33}} & -     & \multicolumn{1}{r|}{-}     & 89.20       & -           & -          \\
Czert                                       & -              & 84.10          & \multicolumn{1}{r|}{-}              & -              & 88.82          & -              & -              & 73.05 & \multicolumn{1}{r|}{-}     & -           & 84.06       & -          \\
Czert + CRF                                 & -              & \underline{\textbf{84.22}} & \multicolumn{1}{r|}{-}              & -              & \textbf{90.29} & -              & -              & 71.36 & \multicolumn{1}{r|}{-}     & -           & 83.70       & -          \\
RuBERT                                      & -              & -              & \multicolumn{1}{r|}{\textbf{62.06}} & -              & -              & 76.97          & -              & -     & \multicolumn{1}{r|}{58.51} & -           & -           & 77.63      \\
RuBERT + CRF                                & -              & -              & \multicolumn{1}{r|}{61.80}          & -              & -              & \textbf{77.69} & -              & -     & \multicolumn{1}{r|}{59.55} & -           & -           & 76.72      \\
\midrule
Slavic-BERT                                 & 79.06          & 78.67          & \multicolumn{1}{r|}{61.42}          & 89.07          & 90.31          & 78.21          & 73.73          & 68.22 & \multicolumn{1}{r|}{59.32} & 83.72       & 78.16       & 77.29      \\
Slavic-BERT + CRF                           & 78.15          & 80.68          & \multicolumn{1}{r|}{63.08}          & 89.97          & 90.13          & 78.72          & 77.76          & 69.12 & \multicolumn{1}{r|}{58.08} & 86.76       & 80.51       & 77.05      \\
XLM-RoBERTa\textsubscript{BASE}                            & 79.53          & 77.89          & \multicolumn{1}{r|}{62.12}          & 88.30          & 89.51          & 77.56          & 76.92          & 68.46 & \multicolumn{1}{r|}{60.45} & 83.25       & 80.89       & 77.21      \\
XLM-RoBERTa\textsubscript{BASE} + CRF                      & 81.10          & 78.80          & \multicolumn{1}{r|}{65.94}          & 88.48          & 90.88          & 77.58          & 79.45          & 73.42 & \multicolumn{1}{r|}{58.86} & 87.02       & 84.20       & 76.87      \\
XLM-RoBERTa\textsubscript{LARGE}                           & 81.43          & 80.58          & \multicolumn{1}{r|}{\underline{\textbf{66.26}}} & \textbf{90.36} & \underline{\textbf{91.62}} & \underline{\textbf{80.22}} & 81.12          & 75.35 & \multicolumn{1}{r|}{61.95} & 87.46       & 86.96       & 77.60      \\
XLM-RoBERTa\textsubscript{LARGE} + CRF                     & \textbf{81.81} & \textbf{81.20} & \multicolumn{1}{r|}{64.95}          & 89.37          & 91.53          & 79.93          & 80.72          & 75.01 & \multicolumn{1}{r|}{61.80} & 86.78       & 87.66       & 77.73     \\
\bottomrule
\end{tabular}%
}

%% file: table_val_lemma.tex
\begin{tabular}{ll|lrr|lrr|lrr}
\toprule
\multicolumn{2}{c|}{\multirow{2}{*}{}}           & \multicolumn{3}{c|}{original data}                                                                         & \multicolumn{3}{c|}{+ PolEval 2019}                                                      & \multicolumn{3}{c}{+ Lexicon}                                                                    \\
\multicolumn{2}{c|}{}                            & \multicolumn{1}{r}{pl}      & cs                                   & ru                                    & \multicolumn{1}{r}{pl}      & cs                          & ru                           & \multicolumn{1}{r}{pl}               & cs                          & ru                          \\
\midrule
\multicolumn{2}{r|}{\textbf{COVID-19}}           & \multicolumn{3}{l|}{\multirow{2}{*}{}}                                                                     & \multicolumn{3}{l|}{\multirow{2}{*}{}}                                                   & \multicolumn{3}{l}{\multirow{2}{*}{}}                                                            \\
\textit{Model}          & \textit{Size}          & \multicolumn{3}{l|}{}       & \multicolumn{3}{l|}{}       & \multicolumn{3}{l}{}                 \\
plT5                    & small                  & \multicolumn{1}{r}{86.36} & -                                    & -                                     & \multicolumn{1}{r}{91.15} & -                           & -                            & \multicolumn{1}{r}{92.02}          & -                           & -                           \\
                        & base                   & \multicolumn{1}{r}{89.99} & -                                    & -                                     & \multicolumn{1}{r}{93.03} & -                           & -                            & \multicolumn{1}{r}{80.70}          & -                           & -                           \\
                        & large                  & \multicolumn{1}{r}{94.05} & -                                    & -                                     & \multicolumn{1}{r}{94.78} & -                           & -                            & \multicolumn{1}{r}{\underline{\textbf{95.36}}} & -                           & -                           \\
mT5                     & small                  & \multicolumn{1}{r}{74.46} & 73.75                              & 70.17                               & \multicolumn{1}{r}{86.80} & 80.98                     & 73.83                      & \multicolumn{1}{r}{81.13}          & 75.45                     & 71.84                     \\
                        & base                   & \multicolumn{1}{r}{87.66} & 85.44                              & 76.96                               & \multicolumn{1}{r}{91.00} & 86.29                     & 76.10                      & \multicolumn{1}{r}{90.42}          & 83.32                     & 75.30                     \\
                        & large                  & \multicolumn{1}{r}{90.57} & 88.84                              & \underline{\textbf{79.09}}                      & \multicolumn{1}{r}{93.76} & \underline{\textbf{89.80}}            & 77.30                      & \multicolumn{1}{r}{93.03}          & 89.27                     & 77.16                     \\
\midrule
\midrule
\multicolumn{2}{r|}{\textbf{USA 2020 Elections}} & \multicolumn{3}{l|}{\multirow{2}{*}{}}                                                                     & \multicolumn{3}{l|}{\multirow{2}{*}{}}                                                   & \multicolumn{3}{l}{\multirow{2}{*}{}}                                                            \\
\textit{Model}          & \textit{Size}          & \multicolumn{3}{l|}{}       & \multicolumn{3}{l|}{}       & \multicolumn{3}{l}{}                 \\
plT5                    & small                  & 83.37                     & -                                    & -                                     & 87.47                     & -                           & -                            & 86.65                              & -                           & -                           \\
                        & base                   & 85.22                     & -                                    & -                                     & 87.89                     & -                           & -                            & 76.80                              & -                           & -                           \\
                        & large                  & 90.97                     & -                                    & -                                     & 90.76                     & -                           & -                            & \underline{\textbf{91.38}}                     & -                           & -                           \\
mT5                     & small                  & 71.46                     & \multicolumn{1}{l}{70.03}          & \multicolumn{1}{l|}{72.18}          & 78.85                     & \multicolumn{1}{l}{75.86} & \multicolumn{1}{l|}{76.18} & 74.54                              & \multicolumn{1}{l}{69.76} & \multicolumn{1}{l}{68.92} \\
                        & base                   & 83.98                     & \multicolumn{1}{l}{80.37}          & \multicolumn{1}{l|}{80.51}          & 84.19                     & \multicolumn{1}{l}{81.97} & \multicolumn{1}{l|}{80.27} & 85.63                              & \multicolumn{1}{l}{78.78} & \multicolumn{1}{l}{78.25} \\
                        & large                  & 88.71                     & \multicolumn{1}{l}{\underline{\textbf{88.33}}} & \multicolumn{1}{l|}{\underline{\textbf{82.86}}} & 89.12                     & \multicolumn{1}{l}{87.27} & \multicolumn{1}{l|}{82.50} & 89.94                              & \multicolumn{1}{l}{86.74} & \multicolumn{1}{l}{81.76} \\
\bottomrule
\end{tabular}

%% file: table_test.tex
\begin{tabular}{l|rrr|rrr}
\toprule
\multicolumn{1}{c|}{\multirow{2}{*}{Submission}} & \multicolumn{3}{c|}{\textbf{Recognition}}        & \multicolumn{3}{c}{\textbf{Normalization}} \\ \cline{2-7} 
                  & pl           & cs           & ru          & pl           & cs           & ru           \\
\midrule
System 1          & 83.33        & 88.08        & 84.30       & 80.27        & 76.62        & 79.32        \\
System 2          & \textbf{85.37}        & \textbf{89.70}        & \textbf{86.16}       & \textbf{82.37}        & \textbf{76.89}        & 81.27        \\
System 3          & 83.40        & 85.19        & 82.77       & 80.32        & 73.06        & \textbf{81.47}        \\
System 4          & 83.33        & 81.70        & 79.20       & 80.27        & 71.11        & 76.84       \\
\bottomrule
\end{tabular}